%% file: paper.tex
\documentclass{article}

\usepackage{microtype}
\usepackage{graphicx}
\usepackage{booktabs} %
\usepackage{xcolor}         %
\usepackage{graphicx}  %
\usepackage{caption}
\usepackage{subcaption} %
\usepackage{lipsum}     %
\usepackage{xspace}
\usepackage{xurl} %

\usepackage{hyperref}
\usepackage{marginnote}

\usepackage[preprint]{icml2026}

\usepackage{amsmath}
\usepackage{amssymb}
\usepackage{mathtools}
\usepackage{amsthm}

\usepackage[capitalize,noabbrev]{cleveref}

\usepackage{wrapfig}  %

\usepackage{booktabs}
\usepackage{tabularx}
\usepackage{multirow}
\usepackage{xcolor}
\usepackage{soul}

\theoremstyle{plain}

\theoremstyle{definition}

\theoremstyle{remark}

\usepackage[textsize=tiny]{todonotes}

\newcommand{\squishlist}{
   \begin{list}{$\bullet$}
    { \setlength{\itemsep}{0pt}      \setlength{\parsep}{3pt}
      \setlength{\topsep}{3pt}       \setlength{\partopsep}{0pt}
      \setlength{\leftmargin}{1.0em} \setlength{\labelwidth}{1em}
      \setlength{\labelsep}{0.5em} } }
\newcommand{\squishend}{
    \end{list}  }
\newcommand{\squishenum}{
   \begin{list}{\arabic{enumi}.}
    { \usecounter{enumi}
      \setlength{\itemsep}{0pt}      \setlength{\parsep}{3pt}
      \setlength{\topsep}{3pt}       \setlength{\partopsep}{0pt}
      \setlength{\leftmargin}{1.0em} \setlength{\labelwidth}{1em}
      \setlength{\labelsep}{0.5em} } }
\newcommand{\squishenumend}{
    \end{list}  }

\newcommand{\name}{FailFast\xspace}

\newcommand{\papertitle}{Fail Fast, Win Big: Rethinking the Drafting Strategy in Speculative Decoding via Diffusion LLMs}

\icmltitlerunning{\papertitle}

\begin{document}

\twocolumn[
  \icmltitle{\papertitle}

\icmlsetsymbol{equal}{*}
\icmlsetsymbol{rui_internship}{*}

\begin{icmlauthorlist}
\icmlauthor{Rui Pan}{princeton,rui_internship}
\icmlauthor{Zhuofu Chen}{princeton}
\icmlauthor{Hongyi Liu}{rice}
\icmlauthor{Arvind Krishnamurthy}{google,uw}
\icmlauthor{Ravi Netravali}{princeton}
\end{icmlauthorlist}

\icmlaffiliation{princeton}{Princeton University}
\icmlaffiliation{uw}{University of Washington}
\icmlaffiliation{google}{Google}
\icmlaffiliation{rice}{Rice University}

\icmlcorrespondingauthor{Rui Pan}{ruipan@princeton.edu}

\icmlkeywords{Machine Learning, ICML}

\vskip 0.3in
]

\printAffiliationsAndNotice{\icmlRuiInternship}

\input{abstract}
\input{intro}

\input{background}

\input{motivation}

\input{method}

\input{evaluation}

\input{conclusion}

\nocite{langley00}

\bibliography{bibliography}
\bibliographystyle{icml2026}

\newpage
\appendix
\onecolumn
\input{appendix}

\end{document}

%% file: abstract.tex
\begin{abstract}

Diffusion Large Language Models (dLLMs) offer fast, parallel token generation, but their standalone use is plagued by an inherent efficiency-quality tradeoff.
We show that, if carefully applied, the attributes of dLLMs can actually be a strength for drafters in speculative decoding with autoregressive (AR) verifiers.
Our core insight is that dLLM's speed from parallel decoding drastically lowers the risk of costly rejections, providing a practical mechanism to effectively realize the (elusive) lengthy drafts that lead to large speedups with speculative decoding.
We present \textbf{\name{}}, a dLLM-based speculative decoding framework that realizes this approach by dynamically adapting its speculation length.
It ``fails fast'' by spending minimal compute in hard-to-speculate regions to shrink speculation latency and ``wins big'' by aggressively extending draft lengths in easier regions to reduce verification latency (in many cases, speculating and accepting 70 tokens at a time!).
Without any fine-tuning, \name{} delivers lossless acceleration of AR LLMs and achieves up to 4.9$\times$ speedup over vanilla decoding, 1.7$\times$ over the best naive dLLM drafter, and 1.7$\times$ over EAGLE-3 across diverse models and workloads.
We open-source \name{} at 
\url{https://github.com/ruipeterpan/failfast}.
\vspace{-5pt}
\end{abstract}

%% file: intro.tex
\vspace{-5pt}
\section{Introduction}
\label{sec:introduction}

\begin{figure*}[t]
    \centering
    \includegraphics[width=0.99\textwidth]{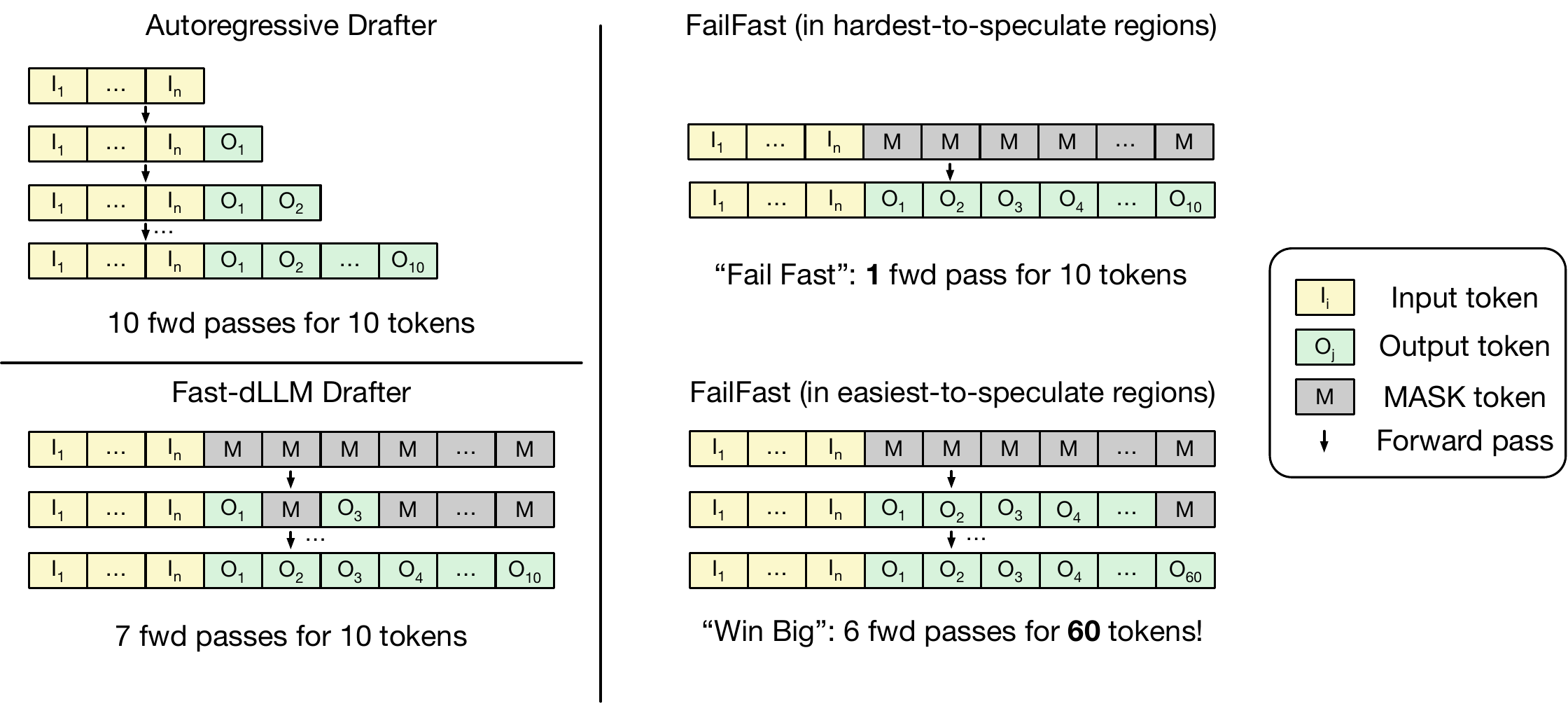}
    \vspace{-5pt}
    \caption{Intuition behind the behavior of \name{} and other baseline drafters in each round of speculation. The AR drafter takes 10 forward passes to speculate 10 tokens, whereas the naive dLLM drafter employs confidence-aware parallel decoding, taking $<10$ forward passes to reach a similar quality. In contrast, \name{} spends minimal compute and dynamically determines how many tokens to propose based on the confidence of speculated tokens. In harder-to-speculate regions (top right), it adopts a shorter speculation length and minimizes the amount of compute to ``fail fast'', further minimizing the speculation latency. In easier-to-speculate regions (bottom right), it aggressively extends the speculation length to ``win big'' and reduce the verification overhead. This example illustrates two extremes of \name{}; in practice, it dynamically navigates through a spectrum of speculation lengths based on decoding difficulty.}
    \label{fig:intuition}
\end{figure*}

A new wave of Diffusion Large Language Models (dLLMs)~\cite{mercury,seed_diffusion,gemini_diffusion,llada2,fast_dllm} has emerged as a compelling alternative to the standard autoregressive paradigm in large language models. Unlike autoregressive (AR) LLMs, which are constrained to generating tokens one by one from left to right, dLLMs possess the unique capability to predict and unmask multiple tokens at arbitrary positions simultaneously. Crucially, this decoding process is highly customizable: the model’s unmasking strategy determines exactly which and how many tokens are unmasked during each denoising step (a model forward pass). As such, dLLMs are highly attractive for low-latency inference.

Yet despite their speed, parallel generation imposes a fundamental limit on modeling accuracy. This limitation stems from the conditional independence assumption required for simultaneous sampling of multiple tokens; by treating tokens generated within the same step as independent of one another, the decoding process inevitably ignores crucial mutual dependencies~\cite{fast_dllm, parallelbench}. Consequently, a direct tension emerges between efficiency and quality. Improving the generation speed (i.e., using fewer forward passes) necessitates unmasking a larger number of tokens per step, which exacerbates the risk of quality degradation. Conversely, maximizing quality forces the sampling procedure to adopt a strict left-to-right, one-token-per-step order that essentially falls back to the speed of autoregressive generation.

While existing work strives to alleviate the stark compute-accuracy tradeoff of dLLMs as standalone generators~\cite{parallelbench,llada2,aup}, this work instead focuses on motivating and realizing a scenario in which we argue that dLLMs are intrinsically beneficial: as draft models in speculative decoding~\cite{spec_decoding} with autoregressive target models. Our proposal extends beyond a simple drop-in replacement of dLLMs as drafters in existing speculative decoding strategies to reap their latency benefits -- indeed, we later show how this can forgo substantial benefits they bring. Instead, our approach is rooted in two key observations that challenge the status quo for both speculative decoding design and considerations around the limitations of dLLMs.

\textbf{First, dLLMs can generate long drafts quickly.} In AR LLM inference, the decoding latency scales with the number of output tokens (i.e., the number of model forward passes)~\cite{decoding_specdecoding,sarathi_serve}, whereas dLLMs can unmask multiple tokens in each forward pass, so the latency is instead linear to the number of model forward passes. dLLMs' ability to generate more tokens quickly motivates a rethink of the central challenge involved in designing effective speculative decoding strategies, i.e., balancing the fact that getting longer drafts accepted yields bigger wins~\cite{turbospec,specd_plus}, but longer drafts come with higher risk due to each token's probability of acceptance dropping exponentially.

\textbf{Second, dLLMs exhibit distinct accuracy-compute concavity patterns at the sequence- and draft-level.}
At the sequence level, the accuracy improvement of each additional forward pass yields diminishing returns. However, within each sequence, there are easier regions of tokens -- e.g., simpler tasks like summarizing prior context~\cite{specreason} -- where minimal compute (i.e., a single forward pass) suffices for accurate generation.
In contrast, for harder regions where existing speculative decoding strategies typically struggle (e.g., difficult tasks like complex arithmetic), dLLMs require more compute to slowly refine their quality. 
This observation drastically relaxes the issues with the latency-accuracy tension inherent to dLLMs: while additional forward passes are necessary for standalone generation, where each token matters for end-to-end quality, they are often unneeded for drafting in speculative decoding. Indeed, beyond the fact that all drafts are ultimately verified (making some errors tolerable), most wins typically come from the easier regions in a sequence~\cite{decoding_specdecoding} where initial dLLM draft accuracy routinely suffices.

Capitalizing on these observations, we present \textbf{\textit{\name{}}}, our dLLM-based speculative decoding framework that revamps the design philosophy of speculative decoding frameworks. Its core operation is governed by two principles:

\squishenum
\item \textbf{Fail Fast:} While most related work attempts to improve the quality of speculated tokens -- e.g., through fine-tuning drafters or adopting an ensemble of small drafters -- the quality of speculation is still fundamentally limited by the capacity of the small drafter model(s). We \emph{fail fast} by deliberately spending minimal compute to speculate on tokens, further reducing drafting latency spent on tokens likely to be rejected anyway, while still (empirically) generating tokens in easy regions with high accuracy.
\item \textbf{Win Big:} In easier regions where the speculated tokens are often accepted (even with minimal drafter compute), we \emph{win big} by aggressively increasing the speculation length -- in many cases, speculating 70 tokens in one round and getting all of them accepted -- to avoid frequently going back and forth between the drafter and verifier, reducing the verification latency. Our signal for speculation easiness is simple and intuitive: the drafter's confidence in its speculated tokens, which we find to be highly correlated with region hardness.
\squishenumend

Across diverse models and workloads, \name{} achieves lossless acceleration of autoregressive target model generation, delivering speedups of up to 4.9$\times$ over vanilla decoding, 1.7$\times$ over a strong dLLM drafter baseline, and 1.7$\times$ over EAGLE-3.
We evaluate on diverse models and datasets and show that the intrinsic properties of dLLMs motivate fundamental changes to drafting strategies, which \name{} exploits to leverage their strengths as draft models.
By enabling off-the-shelf, already-aligned~\cite{fast_dllm_v2, llada2, diffugpt, tiny_a2d} dLLMs to act as effective drafters without any additional training or fine-tuning, \name{} fundamentally changes the drafting paradigm and naturally benefits from continued improvements in dLLM alignment~\cite{specdiff2}.

%% file: background.tex
\section{Background}
\label{sec:background}

\textbf{Diffusion language models.} Recent research has established Diffusion Language Models (dLLMs) as a promising alternative to standard autoregressive generation, with examples spanning both open-source~\cite{llada,llada2,dream} and closed-source models~\cite{mercury, gemini_diffusion, seed_diffusion}. Architecturally, dLLMs retain the standard Transformer backbone, preserving access to familiar metrics such as log-probabilities and token confidence scores. However, they diverge fundamentally in their decoding mechanism: instead of generating tokens strictly left-to-right, dLLMs generate text through an iterative \textit{unmasking} process. This decoding is typically semi-autoregressive~\cite{block_diffusion, gemini_diffusion, fast_dllm_v2, fast_dllm}: the output sequence is divided into blocks, with inter-block attention remaining causal (later blocks attend to earlier blocks), but intra-block attention is bidirectional. (This semi-autoregressive decoding also enables optimizations like KV caching~\cite{fast_dllm}.) Within each block, tokens are unmasked in a non-deterministic order determined by token confidence rather than position, allowing the model to unmask ``easy'' tokens before ``hard'' ones. To further improve efficiency, recent works have proposed acceleration techniques such as approximate KV caching~\cite{fast_dllm, dkv_cache, dllm_cache}, confidence-aware parallel decoding~\cite{fast_dllm_v2, fast_dllm}, and self-speculative decoding~\cite{spiffy}.

\textbf{Speculative decoding.} Borrowing from classical computer architecture principles~\citep{burton1985speculative}, speculative decoding has emerged as a standard technique to alleviate the memory-bound nature~\cite{sarathi_serve} of LLM inference~\citep{stern2018blockwise, spec_decoding, decoding_specdecoding, drafter_selection}. The process operates on a propose-and-verify cycle: in each round, a lightweight drafter first generates a draft, which is subsequently validated by the target model. The speculation phase prioritizes efficiency, relying on methods such as a standalone draft model~\citep{spec_decoding, specinfer}, a trainable module on top of the target model~\citep{medusa, eagle3}, a tree-based token cache~\citep{suffix_decoding, token_recycling, lookahead_alipay}, an n-gram lookup table~\citep{lookahead}, or a retrieval-based datastore~\citep{rest}. In the verification phase, the target model performs a parallel chunked-prefill over these candidates, which usually consists of either a single sequence of tokens as in~\citep{spec_decoding} or tree-like structures to further boost the quality of speculation~\citep{specinfer, medusa, eagle3, sequoia}, and accepts the longest valid prefix. Consequently, the speculation length $n$ is typically conservative (e.g., $n = $ 3-10)~\cite{turbospec,lookahead} to maintain an optimal balance between the speculation overhead and the rate of token acceptance.

%% file: motivation.tex
\section{Motivation}
\label{sec:motivation}

\subsection{dLLMs' Concavity of Accuracy Improvements}

\begin{figure}[t]
    \centering
    \includegraphics[width=0.99\columnwidth]{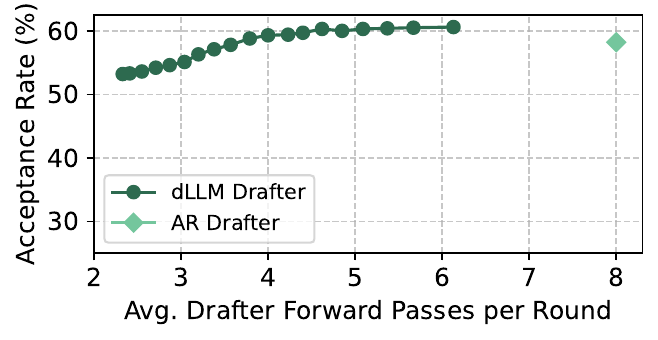}
    \caption{[Qwen2.5-32B, MATH] dLLMs' concavity of accuracy improvements when speculating 8 tokens per round. Doing 2.6$\times$ more drafter forward passes only increased the acceptance rate from 53.2\% to 60.5\%.}
    \label{fig:concavity}
    \vspace{-15pt}
\end{figure}

The quality of a dLLM's output is a function of compute. As non-autoregressive models, dLLMs generate multiple tokens in parallel during an iterative denoising process. In each forward pass, a select few masked tokens with higher confidence within the full sequence of regular and mask tokens are unmasked into regular tokens. As such, the computational cost of a dLLM is predominantly determined by the number of denoising steps (forward passes), not the number of tokens generated. By increasing the number of denoising steps, the model iteratively refines the output, trading latency for higher quality/fidelity.

dLLMs can outperform AR LLMs in generation speed while maintaining accuracy by spending fewer forward passes to generate the same number of output tokens~\cite{fast_dllm}. However, naively deploying them as if they are equally-accurate but faster AR drafters in speculative decoding does not capitalize on their full latency benefit. Crucially, we find that dLLMs show concavity in accuracy wins as more compute is spent -- improvements with each new forward pass have diminishing returns. 
In Fig.~\ref{fig:concavity}, we naively adopt a dLLM as the drafter and show that a higher compute budget leads to diminishing returns in overall acceptance rate.
A dLLM can, in theory, unmask a sequence of unlimited length using a single forward pass. Although the quality of such one-step generation is relatively low and more denoising steps improve the quality -- and those denoising steps are crucial for quality if the dLLM is doing standalone generation -- the role of draft models in speculative decoding is, by definition, to correctly decode the easier tokens, which is a nice match for the fast yet inaccurate one-step generation.

\begin{figure}[t]
    \centering
    \includegraphics[width=0.99\columnwidth]{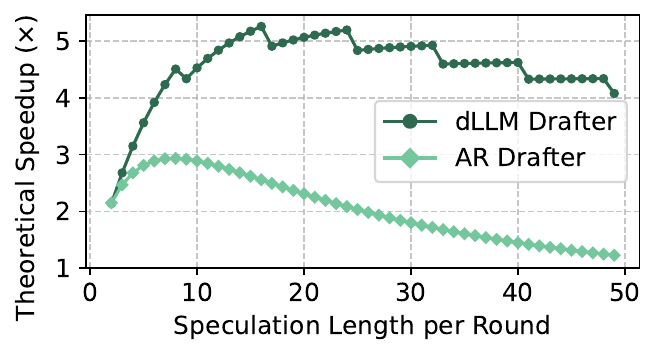}
    \caption{[Qwen2.5-32B, MATH] A semi-autoregressive dLLM that performs one-step generation achieves a higher theoretical speedup than that of an autoregressive drafter.\footnotemark}
    \label{fig:theoretical_analysis}
    \vspace{-15pt}
\end{figure}
\footnotetext{The sawtooth pattern arises because drafting 9–16 tokens costs two forward passes, while 17 tokens requires three (a 50\% latency increase due to semi-autoregressive generation of 8-token blocks).}

In Fig.~\ref{fig:theoretical_analysis}, we analyze the theoretical speedup (Theorem~3.8 in~\cite{spec_decoding}) of speculative decoding over vanilla decoding using different drafters of the same size.
The autoregressive drafter generates one token in each forward pass, incurring drafting latency linear in the draft length, whereas the semi-autoregressive dLLM drafter performs one-step generation and generates 8 tokens in each forward pass~\cite{fast_dllm_v2}, resulting in sublinear drafting latency. 
Even though longer speculation lengths exponentially reduce the acceptance probabilities of the tokens later in the draft, the dLLM's one-step generation compensates by dramatically lowering the average latency to generate each draft token, increasing the maximum speedup from 3.0$\times$ (at speculation length 8) to 5.2$\times$ (at length 16).

\subsection{The Varying Decoding Difficulty within Sequences}

\begin{figure}[t]
    \centering
    \includegraphics[width=0.99\columnwidth]{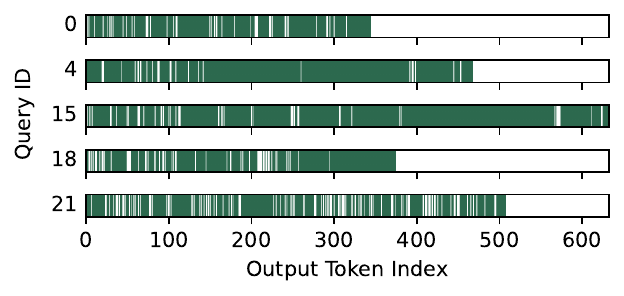}
    \caption{[Qwen2.5-32B, MATH] The varying difficulty of decoding within a sequence. The raster plot visualizes whether each output token is accepted (green, \textcolor[HTML]{2D6A4F}{``easier''}) or rejected and regenerated by the target model (white, ``harder''). Note that an all-white chunk at the right end indicates a shorter sequence.}
    \label{fig:difficulty_intuition}
\end{figure}

\begin{figure}[t]
    \centering
    \includegraphics[width=0.99\columnwidth]{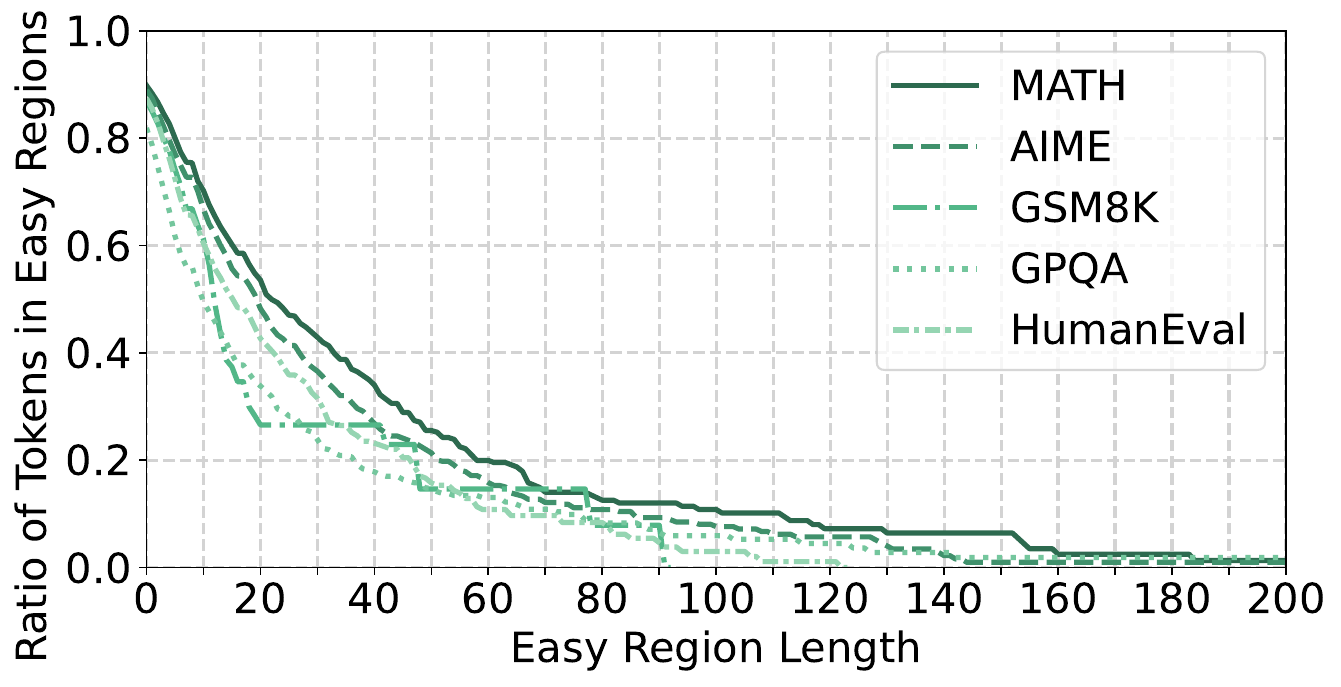}
    \caption{[Qwen2.5-32B] A large fraction of tokens belong to a consecutive block of more than $X$ \textcolor[HTML]{2D6A4F}{``easy''} tokens.}
    \label{fig:consecutive_easy_ratio}
\end{figure}

Importantly, even though the concavity property manifests across entire sequences, within each sequence, there are easier regions and tokens where minimal compute suffices for high-quality generation. 
Similar to prior work~\cite{sd_ee,specd_plus,specreason}, we observe that in natural language generation, the difficulty of generation and speculation differs between regions of the output sequence. Easier regions often consist of syntactic copying, summarization of prior context and input prompt, formulaic enumerations, simple arithmetic, etc. In these regions of lower difficulty, a capable draft model can achieve a near-perfect acceptance rate.
Conversely, harder regions involve complex planning, multi-step reasoning, or knowledge retrieval where the draft model's capacity is insufficient, leading to frequent rejections. In Fig.~\ref{fig:difficulty_intuition}, we pick five random queries from one of our evaluation datasets, run an AR draft model, and classify tokens in the output sequence into \textcolor[HTML]{2D6A4F}{``easier''} tokens (including both correct speculations and the ``bonus tokens'' when all drafted tokens are accepted) and ``harder'' tokens (rejected by the target model).
Further, in Fig.~\ref{fig:consecutive_easy_ratio}, we report the fraction of tokens that belong to a consecutive block of more than $X$ easy tokens across different tasks. If an easy region is defined to be a consecutive block of $>$10 \textcolor[HTML]{2D6A4F}{``easy''} tokens, 50-70\% of tokens belong to an easy region. Moreover, 15-26\% of tokens belong to an easy region of $>$50 tokens.

Standard speculative decoding approaches typically utilize a static speculation length (e.g., $n = 10$ tokens). While this aims to balance speculation overhead and potential speedup, it is suboptimal for the dynamic nature of generation:
\squishlist
    \item \textbf{In \textcolor[HTML]{2D6A4F}{easier} regions} (e.g., tokens 150-380 for query 4 in Fig.~\ref{fig:difficulty_intuition}): A short, fixed speculation length results in undue verification overhead. Even though the drafter could correctly speculate the next 10 tokens, its token generation is forced to pause every 10 tokens to invoke the target model for verification. Since target model inference is memory-bound, the frequent loading of model weights incurs significant latency overhead, preventing the system from reaching peak overall efficiency.
    \item \textbf{In harder regions} (e.g., tokens 200-250 for query 18 in Fig.~\ref{fig:difficulty_intuition}): A fixed speculation length results in undue speculation overhead. The drafter wastes compute generating 10 tokens, most of which are likely to be rejected (after the first or second position), resulting in wasted computation and hurting the end-to-end speedup.
\squishend

Altogether, these properties provide a unique opportunity to change the way drafters are used in speculative decoding.
To cope with the varying difficulty of decoding, an ideal speculative decoding paradigm would dynamically increase the speculation length in easy regions to reduce the frequency of verification and amortize its cost, while minimizing the drafter's compute in hard regions to minimize wasted effort.
Standard autoregressive (AR) draft models are ill-suited for this dynamic strategy -- each AR drafter's forward pass only generates a single token, which makes large speculation lengths (e.g., speculating $>$30 tokens at a time) prohibitively expensive and risky (Fig.~\ref{fig:theoretical_analysis}), since the probability of each token's acceptance drops exponentially, and the drafting latency eventually overshadows the savings from parallel verification.
In contrast, we can reduce the compute budget of dLLM drafters to relax this harsh tradeoff, providing an avenue to get lengthy drafts rapidly.

%% file: method.tex
\section{Method}
\label{sec:method}

Capitalizing on the aforementioned observations, we present \name{}, a speculative decoding framework that employs Diffusion LLMs (dLLMs) as draft models to autoregressive verifiers.
\name{} departs from traditional optimization strategies by \textit{deliberately} minimizing the computational effort of the drafter to \textbf{fail fast} and minimize the speculation latency of the draft model; in regions of lower speculation difficulty, it \textbf{wins big} through aggressively extending the speculation length and reducing the verification latency of the target model.

\subsection{``Fail Fast'': Embracing Error-Proneness for Faster Speculation}
\label{sec:spec_latency}

There has been a myriad of related work on improving the efficacy of speculative decoding by improving the draft acceptance rate. Typically, this is achieved by enhancing the drafted tokens' quality through using an ensemble of small drafters to construct token trees~\cite{specinfer, medusa, eagle3, sequoia}, extensive fine-tuning of the draft model~\cite{online_spec_decoding,atlas,specdiff2}, performing online drafter selection~\cite{drafter_selection}, etc. %

In our work, we propose a counter-intuitive approach: \textbf{we explicitly embrace the error-proneness of drafters in speculative decoding}. 
Instead of refining the draft output to match the target distribution better, we restrict the dLLM to the absolute minimum computational budget -- typically only using one denoising step (model forward pass). Our rationale is twofold:
\squishlist
    \item \textbf{Easy regions:} In regions of low difficulty (e.g., summarization), even a coarse, one-step generation from a small dLLM is often sufficient to speculate the correct tokens (Fig.~\ref{fig:concavity}). Additional refinement steps in these regions are wasted compute.
    \item \textbf{Hard regions:} In regions of high difficulty (e.g., complex reasoning), a small draft model is statistically likely to diverge from the target model regardless of how many refinement steps are applied. Although spending additional compute to refine the draft tokens does indeed improve the acceptance rate, it yields diminishing returns (Fig.~\ref{fig:concavity}): the quality of speculated tokens is still inherently bounded by the capability of the small drafter, and a significant portion of tokens is eventually rejected anyway.
\squishend

By adopting dLLMs as draft models, we relax the requirement for high-quality language modeling compared to using them as standalone models, allowing us to operate in a low-quality but blazing-fast mode that minimizes speculation latency per token.
The verification stage then catches errors, enabling a ``fail fast'' approach that quickly returns the task of decoding to the target model with minimal speculation overhead when the task becomes difficult.

\subsection{``Win Big'': Dynamic Expansion of Speculation Length for Faster Verification}
\label{sec:verif_latency}

While reducing speculation latency addresses the cost of generating drafts, it does not lower the verification latency -- the overhead incurred by executing the large target model, which is a factor of both the number of verifications (rounds of speculation) and the verification latency in each round. If we strictly use short speculation lengths, the target model is invoked frequently, creating a latency bottleneck.

To mitigate this, \name{} capitalizes on ``easy'' segments by opportunistically increasing the speculation length. 
When the draft model encounters a subsequence that is easier to speculate, we defer verification and aggressively extend the length of the speculated draft.
Because speculation has, by design, such a low cost in \name{}, the risk of ``failing'' is low, and we can afford these aggressive speculations.

\paragraph{Confidence as a difficulty signal.} 
To detect these easy regions without ground-truth knowledge of difficulty,  we utilize the dLLM's internal confidence~\cite{deepconf,apparate} as a proxy for speculation difficulty. Although dLLMs are non-autoregressive, they use the same Transformers backbone as AR LLMs, so we still have access to the probability distribution over the vocabulary for each token. For a dLLM predicting a token $x_i$ at position $i$, we define its confidence $C_i$ as the maximum probability in the output distribution $P_i$: 
$C_i = \max_{v \in V} P_i(x_i = v)$, where $V$ is the vocabulary.

\begin{algorithm}[tb]
  \caption{\name{} Main Logic in Each Round}
  \label{alg:adaptive_speculation}
  \begin{algorithmic}
    \STATE {\bfseries Hyperparameters:} step size $N$, confidence threshold $\tau$, max length $N_{\max}$
    \STATE $L \leftarrow 0$ \COMMENT{Total number of speculated tokens in this round}

    \WHILE{true}
      \STATE $L \leftarrow L + N$
      \STATE Speculate the next $N$ tokens.

      \IF{any speculated token has confidence  $C_i < \tau$ \\ \textbf{or} $L \ge N_{\max}$}
        \STATE \textbf{break}
      \ENDIF
    \ENDWHILE

    \STATE Submit all $L$ tokens for verification.
  \end{algorithmic}
\end{algorithm}
\vspace{-5pt}

\paragraph{Dynamic speculation length expansion logic.}
Our adaptive strategy is detailed in Algorithm~\ref{alg:adaptive_speculation}. We begin with a default speculation length $N$. After generating these initial tokens (using minimal compute), we inspect their confidence scores.
If all tokens' confidence in the current speculated sequence exceeds a confidence threshold $\tau$ (indicating an ``easy'' region), we assume the draft will likely be accepted. Instead of stopping to verify, we immediately extend the speculation length by another $N$ tokens and invoke the drafter again.
We repeat this extension process until a low-confidence token is detected or a maximum sequence length $N_{\max}$ is reached.
This mechanism allows \name{} to generate massive chunks of tokens (e.g., 70 tokens) in easy regions in a single round of speculation, and to get a majority of them accepted in many cases, which drastically reduces the number of times the target model needs to be invoked (verification rounds) -- the verification overhead in each round stays the same because short prefills are memory-bound~\cite{sarathi_serve} -- while reverting to short bursts in harder regions.

While the notion of dynamically adjusting the draft length is an established concept in speculative decoding~\cite{turbospec}, the resulting speedups are often limited, hindered by the risk of invoking undue speculation latency.
For example, SpecDec++~\cite{specd_plus} achieves an average speedup of $\sim$10\% over its baselines.
In comparison, \name{} overcomes this by using minimal compute to minimize the risk of incorrect speculation, allowing it to drastically adapt the speculation length on-the-fly.

We include a discussion of alternative design choices and optimizations in Appendix~\ref{sec:failed_explorations}.

\subsection{Policy Design vs. Training-Based Approaches}

Our work focuses specifically on how to adapt drafting policies when using dLLMs as drafters, rather than improving performance through additional training.
Indeed, most state-of-the-art dLLMs are adapted from pretrained autoregressive (AR) weights~\cite{fast_dllm_v2, llada2, diffugpt, tiny_a2d} and are therefore already implicitly aligned with the target AR models when used as drafters. 
Leveraging this property, we study how to better exploit dLLMs at inference time via policy design. 
That said, our approach is complementary to prior optimizations that aim to improve the overall acceptance rate of drafted tokens~\cite{specinfer, sequoia, online_spec_decoding, atlas, drafter_selection}, including recent work that fine-tunes the dLLM drafter for improved alignment and higher acceptance rates~\cite{specdiff,specdiff2}. These techniques are \textit{complementary} to \name{}, as the dLLMs we employ can be further fine-tuned to fit the base model's distribution better.

%% file: evaluation.tex
\section{Evaluation}
\label{sec:evaluation}

\begin{table*}[ht]
\centering
\caption{Comparison of speedups of different drafting methods over vanilla decoding. \name{} achieves the \textbf{highest} average speedup.}
\label{tab:e2e_speedup}

\small 
\setlength{\tabcolsep}{8pt} 

\begin{tabular}{lcccccc}
\toprule
\multirow{2}{*}{\textbf{Drafting Method}} & \multicolumn{5}{c}{\textbf{Dataset}} \\
\cmidrule(lr){2-6}
& \textbf{MATH} & \textbf{AIME} & \textbf{GSM8K} & \textbf{GPQA} & \textbf{HumanEval} & \textbf{Average} \\
\midrule
\multicolumn{7}{c}{\textbf{Target Model: Qwen2.5-32B-Instruct}} \\ 
\midrule
AR Draft Model
& $ {2.93\times}$ & $ {2.84\times}$ & $ {2.82\times}$ & $ {2.26\times}$ & $ {2.72\times}$ & $ {2.71\times}$ \\ 

\midrule
Fast-dLLM
& $ {3.57\times} $ & $ {3.29\times}$ & $ {3.10\times}$ & $ {2.42\times}$ & $ {3.16\times}$ & $ {3.11\times}$ \\

\midrule
EAGLE-3 (w/ draft tree)
& $ {2.87\times}$ & $ {2.85\times}$ & $ {2.59\times}$ & $ {2.42\times}$ & $ {3.31\times}$ & $ {2.81\times}$ \\ 

\midrule
\name{} (ours)
& $ {\mathbf{4.90\times}}$ & $ {\mathbf{4.40\times}}$ & $ {\mathbf{3.71\times}}$ & $ {\mathbf{3.11\times}}$ & $ {\mathbf{4.06\times}}$ & $ {\mathbf{4.04\times}}$ \\

\midrule 
\multicolumn{7}{c}{\textbf{Target Model: Qwen2.5-14B-Instruct}} \\ 
\midrule
AR Draft Model
& $ {2.05\times}$ & $ {1.94\times}$ & $ {1.95\times}$ & $ {1.71\times}$ & $ {1.91\times}$ & $ {1.91\times}$ \\ 

\midrule
Fast-dLLM
& $ {2.57\times}$ & $ {2.22\times}$ & $ {2.22\times}$ & $ {1.86\times}$ & $ {2.23\times}$ & $ {2.22\times}$ \\

\midrule
EAGLE-3 (w/ draft tree)
& $ {2.48\times}$ & $ {2.38\times}$ & $ {2.23\times}$ & $ {2.04\times}$ & $ {2.68\times}$ & $ {2.36\times}$ \\ 

\midrule
\name{} (ours)
& $ {\mathbf{3.92\times}}$ & $ {\mathbf{3.37\times}}$ & $ {\mathbf{3.04\times}}$ & $ {\mathbf{2.54\times}}$ & $ {\mathbf{3.41\times}}$ & $ {\mathbf{3.26\times}}$ \\ 

\midrule 
\multicolumn{7}{c}{\textbf{Target Model: Qwen2.5-7B-Instruct}} \\ 
\midrule
AR Draft Model
& $ {1.43\times}$ & $ {1.42\times}$ & $ {1.41\times}$ & $ {1.25\times}$ & $ {1.40\times}$ & $ {1.38\times}$ \\ 

\midrule
Fast-dLLM
& $ {1.95\times}$ & $ {1.57\times}$ & $ {1.54\times}$ & $ {1.34\times}$ & $ {1.61\times}$ & $ {1.60\times}$ \\

\midrule
EAGLE-3 (w/ draft tree)
& $ {2.31\times}$ & $ {2.36\times}$ & $ {2.02\times}$ & $ {\mathbf{1.98\times}}$ & $ {2.39\times}$ & $ {2.21\times}$ \\ 

\midrule
\name{} (ours)
& $ {\mathbf{3.06\times}}$ & $ {\mathbf{2.63\times}}$ & $ {\mathbf{2.34\times}}$ & $ {1.89\times}$ & $ {\mathbf{2.71\times}}$ & $ {\mathbf{2.52\times}}$ \\  

\bottomrule

\end{tabular}

\end{table*}

\subsection{Setup}

\textbf{Models and Baselines.} In our main experiments, we evaluate three target models of different sizes: \texttt{Qwen2.5-\{32B,14B,7B\}-Instruct}~\citep{qwen2.5}. The different speculative decoding schemes we evaluate are listed below. For all schemes, we set the generation temperature of all models to 0 and set the maximum number of output tokens to 1024.
We discuss more details of our best-effort reproduction of all baselines and additional baselines in Appendix~\ref{sec:baselines}. 

\squishlist
    \item \textbf{AR drafter~\cite{spec_decoding}:} We adopt \texttt{Qwen2.5-1.5B-Instruct} as our autoregressive speculative sampling baseline, which uses the same training data as Qwen2.5-32B. For each dataset, we conduct an extensive parameter sweep ($n = $ 3-20) and pick the speculation length that achieves the best speedup.
    \item \textbf{Fast-dLLM~\cite{fast_dllm_v2}:} For the dLLM drafter baseline, we employ \texttt{Fast\_dLLM\_v2\_1.5B}~\cite{fast_dllm_v2}, the state-of-the-art semi-autoregressive diffusion LLM that embeds efficiency optimizations such as approximate caching and confidence-aware parallel decoding. We adopt its default hyperparameters that achieve the best balance between accuracy and generation speed. This baseline represents naively plugging in a dLLM as the draft model without the additional optimizations of \name{}.
    \item \textbf{EAGLE-3~\cite{eagle3}:} 
    EAGLE-3 is one of the most widely deployed state-of-the-art speculative decoding drafters. As a lightweight single-layer autoregressive Transformer, EAGLE-3 introduces a multi-layer feature fusion and a training-time test mechanism for improving drafting quality. Different from other strategies where a single chain of draft tokens is proposed for verification, EAGLE-3 leverages a draft-tree-based mechanism that improves the overall token acceptance rate.
    \item \textbf{\name{}:} \name{} employs \texttt{Fast\_dLLM\_v2\_1.5B} as the underlying model and performs one-step generation and dynamic speculation length expansion. For the hyperparameters in Alg.~\ref{alg:adaptive_speculation}, we set $\tau=\{0.4,0.45,0.5\}$ for the \{32,14,7\}B target model and $N=10$. Although picking specific hyperparameters for each dataset/model combination yields higher speedups, we use the same set of hyperparameters across datasets to demonstrate generalizability. \name{}'s performance degrades gracefully as hyperparameters are tweaked (Appendix~\ref{sec:hyperparam_sensitivity}).
\squishend

\textbf{Datasets.} Similar to prior work~\cite{fast_dllm_v2,specdiff2,eagle3}, we evaluate \name{} on a wide range of diverse benchmarks: MATH, AIME, and GSM8K~\cite{math500,aime,gsm8k} for mathematical reasoning, GPQA~\cite{gpqa} for knowledge-intensive question answering, and HumanEval~\cite{humaneval} for code generation.

\textbf{Hardware.} We run our evaluations on two NVIDIA A6000-48GB GPUs connected via PCIe 4.0 x16. We profile latency using SGLang~\cite{sglang}, enable prefix caching~\citep{marconi} for both the draft model and target model, and use TP=2 for all target models. %

\vspace{-2pt}

\subsection{End-to-End Results}

Table~\ref{tab:e2e_speedup} presents the end-to-end speedups over vanilla decoding across different schemes and datasets. 

\textbf{Workload sensitivity.} Across models and datasets, \name{} achieves an overall speedup of 1.9-4.9$\times$. Compared to baselines, \name{} achieves a speedup of 1.3-2.1$\times$ over the AR drafter baseline and a speedup of 1.2-1.7$\times$ over the Fast-dLLM baseline. \name{} has a higher win over baselines on datasets that are easier to speculate: math and coding workloads exhibit centralized regions of predictable tokens (e.g., intermediate calculations and code generation), allowing \name{} to capitalize on long speculation windows. Conversely, datasets where ``easy'' tokens are scattered rather than clustered offer fewer opportunities for aggressive speculation length expansion, though \name{} still outperforms fixed-length baselines with moderate length expansion.

\begin{figure}[t]
    \centering
    \includegraphics[width=0.99\columnwidth]{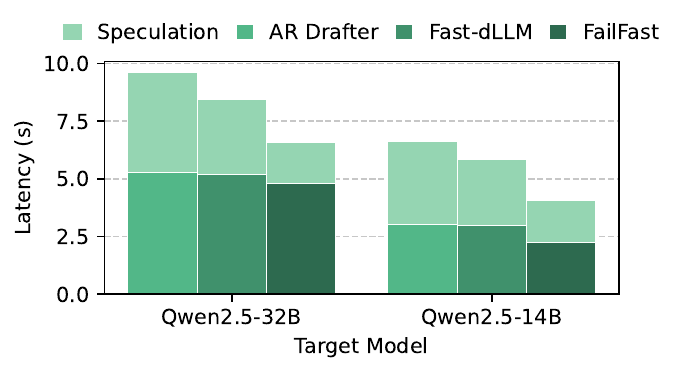}
    \caption{Breakdown of end-to-end latency into speculation (lighter, upper) and verification (darker, lower). While applying a dLLM drafter (Fast-dLLM) reduces the speculation latency, \name{} further reduces the speculation latency (in each round) while also reducing the verification latency (the number of rounds).}
    \label{fig:latency_breakdown}
\end{figure}

\begin{figure}[t]
    \centering
    \includegraphics[width=0.99\columnwidth]{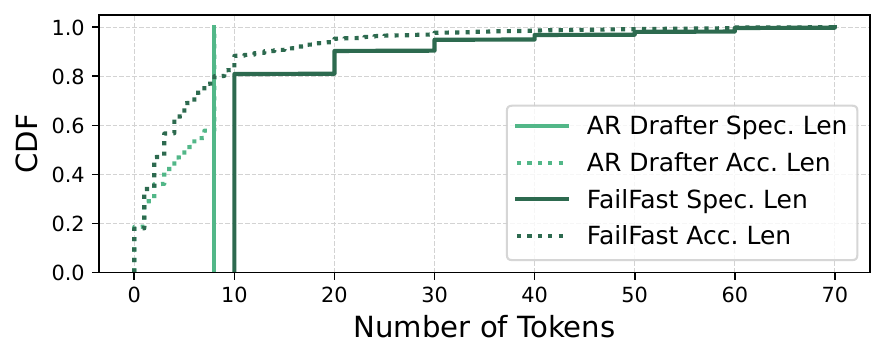}
    \caption{[Qwen2.5-32B, MATH] CDF of the number of accepted/speculated tokens in each round. Full results are in Tab.~\ref{tab:performance_details}.}
    \label{fig:cdf}
\end{figure}

\textbf{Efficiency of \name{}.} Naively plugging in a dLLM as the drafter (Fast-dLLM) retains the acceptance rate (Tab.~\ref{tab:performance_details}) and achieves a speedup over the AR drafter. However, as shown in Fig.~\ref{fig:latency_breakdown}, this gain is primarily derived from a reduction in speculation latency (22.9\% on average)\footnote{This improvement falls short of the theoretical end-to-end speedups of dLLMs~\cite{fast_dllm_v2} because our termination criterion requires the next continuous $n$ tokens to be unmasked; due to the random decoding order within token blocks, the leftmost tokens are not necessarily decoded first.} while the verification latency remains near-constant.
In contrast, \name{} further reduces speculation latency significantly (a further 41.1\% over Fast-dLLM) by employing an ultra-small compute budget for the drafter. Although our approach inevitably degrades the acceptance rate (Tab.~\ref{tab:performance_details}) -- a combined result of a low drafter compute budget and proposing more tokens on average in each round -- we achieve superior end-to-end speedups (e.g., 3.1-4.9$\times$ for Qwen2.5-32B). In addition, \name{} reduces verification latency (Fig.~\ref{fig:latency_breakdown}) by an average of 17.1\% via adopting adaptive speculation lengths: through dynamically expanding the speculation window in ``easy'' regions, we can speculate and verify up to 70 tokens (Fig.~\ref{fig:cdf}) in a single round. This drastic reduction in the total number of speculation-verification rounds and the speculation overhead in each round outweighs the cost of a lower per-token acceptance rate. We include an example trajectory in Appendix~\ref{sec:traj_visualization} to visualize \name{} in action.

\textbf{Detailed analyses.} Due to space constraints, we only further dissect \name{}'s performance improvements on Qwen2.5-32B on the MATH dataset, and we report the data on all other datasets and model combinations in Appendix~\ref{sec:detailed_perf_comparison}.
Even though the AR drafter has a higher overall acceptance rate (Tab.~\ref{tab:performance_details}), for each of its forward passes, it produces only one token and accepts 0.6 tokens on average. In contrast, \name{} can produce an average of 4.1 tokens per forward pass and accept 1.7 of them.
Another source of \name{}'s win comes from its aggressive speculation length expansion. Fig.~\ref{fig:cdf} shows a CDF of the accepted/speculated length across rounds. In $\sim$20\% of the rounds, we extended the speculation length beyond the default, reducing the average number of speculation-verification rounds by 16.0\%.
Interestingly, \name{}’s dynamic policy converges to an average speculation length of 14.6 tokens, closely matching the theoretically optimal of 16 tokens from our prior analysis (Fig.~\ref{fig:theoretical_analysis}).

\textbf{Comparisons with EAGLE-3.}
Across models, \name{} achieves an average speedup of 1.0–1.4$\times$ over EAGLE-3, with gains of up to 1.3–1.7$\times$ on specific datasets.
Even though EAGLE-3 incurs a smaller drafting latency in each of its forward passes -- it only has one layer compared to the 28 layers in other drafters, reducing overheads associated with loading weights and launching kernels -- \name{} still outperforms EAGLE-3.
This is because EAGLE-3 remains an autoregressive drafter that generates only one token per forward pass, whereas \name{} generates multiple tokens in each pass.
Moreover, while EAGLE-3 speculates a tree of 50–60 tokens per round, its draft tree is relatively shallow (default depth 8~\cite{eagle3}) compared to \name{}, which drafts a chain of up to 70 tokens in a single round.
As a result, EAGLE-3’s upper bound on the number of accepted tokens is constrained by tree depth, limiting its ability to exploit extended regions of easier tokens.

Due to space constraints, we refer readers to the Appendix sections for a more in-depth discussion regarding EAGLE-3, the implementation details of baselines, a hyperparameter sensitivity analysis of \name{}, and a discussion of explorations of alternative design choices.

%% file: conclusion.tex
\section{Conclusion}

In this paper, we present \name{}, a novel speculative decoding framework that leverages dLLMs to achieve significant lossless acceleration. By spending minimal compute in hard regions to ``fail fast'' and aggressively extending speculation lengths in easier segments to ``win big,'' \name{} minimizes both speculation and verification latencies. Without requiring fine-tuning, the framework achieves up to 4.9$\times$ speedup over vanilla decoding and significantly outperforms state-of-the-art baselines like Fast-dLLM and EAGLE-3, demonstrating that dLLMs are uniquely suited for adaptive, high-efficiency drafting.

\textbf{Impact Statement} This paper presents work whose goal is to advance the field of machine learning. There are many potential societal consequences of our work, none of which we feel must be specifically highlighted here.

\textbf{Acknowledgments} We thank Princeton's Systems for Artificial Intelligence Lab (SAIL) and Princeton Language and Intelligence (PLI) for providing the hardware resources for running experiments.
Rui would like to thank the many full-time researchers and interns at Google SRG for their help and support, especially Nikos Pagonas, Howie Chen, Manos Chatzakis, Vinay Banakar, Liana Patel, Jiani Zhang, and Lily Tsai. 
In addition, Rui would like to thank Minghao Yan for a random conversation on speculative decoding back in 2024 that planted the seed of this work and for providing feedback on an earlier version of this paper.

%% file: appendix.tex
\section{List of abbreviations and notations}
\begin{table}[h]
    \centering
    \caption{Glossary of abbreviations, notations, and terminology.}
    \label{tab:list_of_notations}
    \begin{tabular}{cc}
        \toprule
        \textbf{Notation} & \textbf{Definition} \\
        \midrule
        AR & Autoregressive \\
        dLLM & Diffusion language models \\
        Round & A speculation-verification round \\
        $n$ / $N$ & Speculation length (number of tokens) in each round \\

        \bottomrule
    \end{tabular}
\end{table}

\section{Detailed Performance Comparison}
\label{sec:detailed_perf_comparison}

\begin{table*}[h!]
\centering
\caption{Detailed performance comparison. We report the average acceptance rate, average accepted/speculated lengths in each round\footnotemark, maximum accepted/speculated lengths across rounds, average number of speculation-verification rounds, and the average number of drafter forward passes in each round.}
\label{tab:performance_details}

\footnotesize 
\setlength{\tabcolsep}{3.5pt} %

\begin{tabular}{llcccccc}
\toprule
\multirow{2}{*}{\textbf{Drafting Method}} & \multirow{2}{*}{\textbf{Metric}} & \multicolumn{5}{c}{\textbf{Dataset}} \\
\cmidrule(lr){3-7}
& & \textbf{MATH} & \textbf{AIME} & \textbf{GSM8K} & \textbf{GPQA} & \textbf{HumanEval} & \textbf{Average} \\
\midrule

\multicolumn{8}{c}{\textbf{Target: Qwen2.5-32B-Instruct}} \\
\midrule

\multirow{5}{*}{\textbf{AR drafter}} 
    & Acceptance Rate           & 58.2\%   & 56.0\%   & 55.3\%   & 51.2\% & 56.1\% & 55.4\% \\
    & Avg (Acc./Spec.) Len  & 4.7 / 8  & 4.5 / 8  & 4.4 / 8 & 2.6 / 5 & 3.9 / 7 & 4.0 / 7.2 \\
    & Max (Acc./Spec.) Len  & 8 / 8 & 8 / 8 & 8 / 8 
    & 5 / 5 & 7 / 7 & - \\
    & Num Speculation Rounds        & 93.6     & 138.8    & 56.1    & 138.3 & 76.6 & 100.7 \\
    & Avg Drafter Passes / Round  & 8      & 8      & 8      & 5 & 7 & 7.2 \\
\midrule

\multirow{5}{*}{\textbf{Fast-dLLM}} 
    & Acceptance Rate           & 60.5\%   & 56.9\%   & 53.2\%   & 52.7\% & 57.7\% & 56.2\% \\
    & Avg (Acc./Spec.) Len  & 4.8 / 8  & 4.5 / 8  & 4.3 / 8 & 2.6 / 5 & 4.0 / 7 & 4.0 / 7.2  \\
    & Max (Acc./Spec.) Len  & 8 / 8 & 8 / 8 & 8 / 8 & 5 / 5 & 7 / 7 & - \\
    & Num Speculation Rounds        & 90.4     & 137.4    & 57.3    & 135.3 & 74.3 & 98.9 \\
    & Avg Drafter Passes / Round  & 5.7      & 6.1      &  6.1     & 4.4 & 5.2 & 5.5 \\
\midrule

\multirow{5}{*}{\textbf{\name{}}} 
    & Acceptance Rate           & 40.6\%      & 38.0\%       & 33.6\%    & 26.4\% & 31.7\% & 34.1\% \\
    & Avg (Acc./Spec.) Len  & 6.0 / 14.6  & 5.1 / 13.3   & 4.1 / 12.1  & 3.3 / 12.0 & 4.5 / 11.3 & 4.6 / 12.7  \\
    & Max (Acc./Spec.) Len  & 70 / 70   & 70 / 70    & 50 / 70 & 70 / 70 & 60 / 70 & - \\
    & Num Speculation Rounds        & 78.6        & 129.9    &  59.1  & 119.0 & 66.9 & 90.7 \\
    & Avg Drafter Passes / Round  & 3.6         & 3.4          &  3.2 & 3.1 & 3.1 & 3.3 \\

\midrule
\multicolumn{8}{c}{\textbf{Target: Qwen2.5-14B-Instruct}} \\
\midrule

\multirow{5}{*}{\textbf{AR drafter}} 
    & Acceptance Rate           & 68.3\%   & 66.8\%   & 64.1\%   & 60.2\% & 69.9\% & 65.9\% \\
    & Avg (Acc./Spec.) Len  & 4.1 / 6  & 3.3 / 5  & 3.8 / 6 & 2.4 / 4 & 2.8 / 4 & 3.3 / 5.0 \\
    & Max (Acc./Spec.) Len  & 6 / 6 & 5 / 5 & 6 / 6 & 4 / 4 & 4 / 4 & - \\
    & Num Speculation Rounds        & 93.0       & 175.5    & 75.5 & 176.3 & 89.3 & 121.9 \\
    & Avg Drafter Passes / Round  & 6      & 5      &  6      & 4 & 4 & 5 \\
\midrule

\multirow{5}{*}{\textbf{Fast-dLLM}} 
    & Acceptance Rate           & 68.9\%   & 66.8\%   & 61.9\%   & 61.3\% & 73.9\% & 66.6\% \\
    & Avg (Acc./Spec.) Len  & 4.1 / 6  & 3.3 / 5  & 3.7 / 6 & 2.5 / 4 & 3.0 / 4 & 3.3 / 5.0 \\
    & Max (Acc./Spec.) Len  & 6 / 6 & 5 / 5 & 6 / 6 & 4 / 4 & 4 / 4 & - \\
    & Num Speculation Rounds        & 93.0       & 175.3    & 78.1 & 173.8 & 86.6 & 121.4  \\
    & Avg Drafter Passes / Round  & 4.2      & 4.0      & 4.6      & 3.6 & 3.2 & 3.9 \\
\midrule

\multirow{5}{*}{\textbf{\name{}}} 
    & Acceptance Rate           & 43.5\%      & 39.2\%       & 36.4\%      & 29.2\% & 39.3\% &  37.5\% \\
    & Avg (Acc./Spec.) Len  & 6.5 / 14.5  & 5.2 / 12.7   & 4.4 / 12.2  & 3.5 / 11.7 & 5.2 / 13.2 & 5.0 / 12.9 \\
    & Max (Acc./Spec.) Len  & 60 / 70   & 70 / 70    &  60 / 60  & 70 / 70 & 60 / 70 & - \\
    & Num Speculation Rounds        & 71.4        & 127.8        &  67.1 & 136.1 & 55.4 & 91.6 \\
    & Avg Drafter Passes / Round  & 3.6         & 3.2          & 3.2 & 3.0 & 3.3 & 3.3 \\

\bottomrule
\end{tabular}
\end{table*}

\footnotetext{The acceptance rate does not exactly equal the average accepted length divided by the average speculation length, because the average of averages does not necessarily equal the overall average. Note that, unlike prior work~\cite{specdiff2}, we \textbf{do not} add one to the average accepted length to represent the token (either correcting the rejected token, or the bonus token if none is rejected) from the verifier.}

\section{Hyperparameter Sensitivity Analysis of \name{}}
\label{sec:hyperparam_sensitivity}

\begin{figure*}[h]
    \centering
    \begin{subfigure}{0.48\textwidth}
        \centering
        \includegraphics[width=\linewidth]{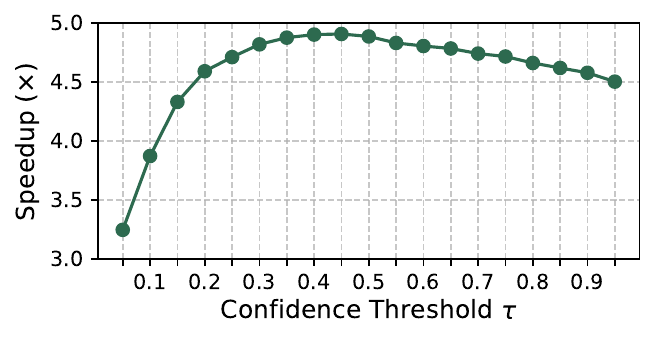}
        \caption{Impact of the confidence threshold $\tau$ in Alg.~\ref{alg:adaptive_speculation}.}
        \label{fig:micro_threshold}
    \end{subfigure}
    \hfill %
    \begin{subfigure}{0.48\textwidth}
        \centering
        \includegraphics[width=\linewidth]{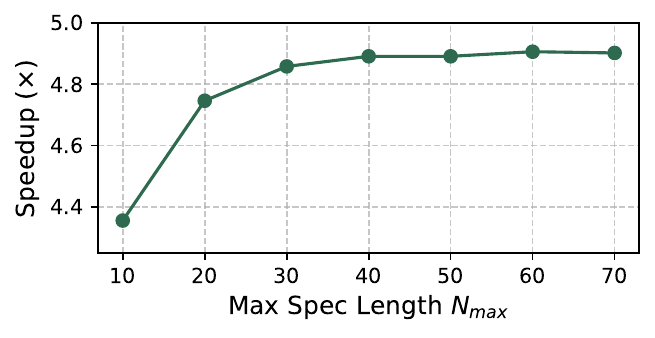} %
        \caption{Impact of the maximum speculation length $N_{max}$ in Alg.~\ref{alg:adaptive_speculation}.}
        \label{fig:micro_maxlen}
    \end{subfigure}
    
    \caption{Impact of \name{}'s hyperparameters on its performance. \name{}'s performance degrades gracefully as the hyperparameters are tweaked.}
    \label{fig:cdf}
\end{figure*}

We analyze the sensitivity of \name{} to its key hyperparameters to understand the trade-offs between speculation overhead and verification efficiency. We focus our analysis on Qwen2.5-32B on MATH, and vary its default hyperparameters $\tau=0.4$ and $N=10$.
One advantage of \name{} is that, because drafting is low-latency by design, it does not require extensive tuning of hyperparameters to reach optimal performance, and the speculation length is self-adaptive based on token difficulty,
whereas traditional speculative decoding schemes typically require workload-aware tuning of the speculation length $N$ to reach optimal performance.

\textbf{Confidence Threshold ($\tau$).} The confidence threshold controls the aggressiveness of the speculation length expansion. We find that $\tau \in [0.3, 0.55]$ yields optimal performance. A threshold that is too high (e.g., 0.7) makes \name{} overly conservative, preventing the proposal of long sequences in easy regions. Conversely, a threshold that is too low (e.g., 0.1) increases speculation latency overhead by generating tokens that are statistically likely to be rejected by the verifier.

\textbf{Maximum Speculation Length ($N_{\max}$).} We observe that $N_{\max}$ values between 30 and 70 perform reliably well. We adopt a default value of $N_{max}=60$ because LLM prefill for short sequences is typically memory-bandwidth bound, and the typical tipping point between memory-bound and compute-bound for medium-sized LLMs typically ranges between 64 and 128 tokens on modern GPUs~\cite{sarathi_serve}. In high-throughput inference settings where batch sizes are large, bigger values of $N_{\max}$ still work, although they might inflate the verification latency slightly -- the prefill pass would eventually become compute-bound for large sequence lengths, even with prefix caching~\cite{marconi}. In that case, synchronous verification might force a reduction in $N_{\max}$ to avoid inflating the per-round verification latency, and \name{}'s performance improvement from ``winning big'' might drop slightly. However, \name{}'s performance improvement from ``failing fast'' persists. 
We note that speculative decoding is inherently more suitable for inference acceleration at smaller batch sizes to begin with and provides diminishing gains as batch size increases, where target model inference becomes less memory-bound and more compute-bound~\cite{magicdec,amd_sd_blog}.

\section{Best-Effort Baselines}
\label{sec:baselines}

We evaluate \name{} against a diverse set of strong baselines, covering both single-layer and multi-layer drafters and single-token predictors and multi-token predictors (in each drafter forward pass). We acknowledge that there exist many other speculative decoding methods, e.g., Medusa~\cite{medusa}, Lookahead Decoding~\cite{lookahead}, etc. Since none of them have been incorporated into high-throughput inference engines~\cite{vllm,sglang} and EAGLE-3 outperforms them~\cite{eagle3, specbench}, we omit those schemes from our evaluations. 

\subsection{Fast-dLLM Drafter}

Our Fast-dLLM baseline adopts the same speculation length as the AR drafter. To show the full potential of a naive baseline, we also conduct an extensive parameter sweep ($n = $ 3-20) for the optimal speculation length of the dLLM drafter and pick a (different) length that achieves the best speedup for each dataset. The end-to-end speedup is reported in Tab.~\ref{tab:fastdllm_vs_fastdllm_plus}. This baseline is idealistic because it requires extensive offline profiling to find the optimal tradeoff between the acceptance rate and speculation latency, which is indeterministic because of Fast-dLLM's confidence-aware parallel decoding. More importantly, we find that on average, picking the best speculation length only brings a 2.3-3.3\% performance improvement over using the same speculation length as the AR drafter, so we omit Fast-dLLM+ from the main paper for conciseness.

\begin{table*}[ht]
\centering
\caption{Comparison of Fast-dLLM (same speculation length as the AR drafter) with Fast-dLLM+ (best speculation length on each dataset from a parameter sweep).}
\label{tab:fastdllm_vs_fastdllm_plus}

\small 
\setlength{\tabcolsep}{8pt} 

\begin{tabular}{lcccccc}
\toprule
\multirow{2}{*}{\textbf{Drafting Method}} & \multicolumn{5}{c}{\textbf{Dataset}} \\
\cmidrule(lr){2-7}
& \textbf{MATH} & \textbf{AIME} & \textbf{GSM8K} & \textbf{GPQA} & \textbf{HumanEval} & \textbf{Average} \\
\midrule
\multicolumn{7}{c}{\textbf{Target Model: Qwen2.5-32B-Instruct}} \\ 
\midrule

Fast-dLLM
& $ {3.57\times} $ & $ {3.29\times}$ & $ {3.10\times}$ & $ {2.42\times}$ & $ {3.16\times}$ & $ {3.11\times}$ \\

\midrule
Fast-dLLM+
& $ {3.75\times}$ & $ {3.34\times}$ & $ {3.10\times}$ & $ {2.48\times}$ & $ {3.23\times}$ & $ {3.18\times}$ \\

\midrule

\multicolumn{7}{c}{\textbf{Target Model: Qwen2.5-14B-Instruct}} \\ 
\midrule

Fast-dLLM
& $ {2.57\times}$ & $ {2.22\times}$ & $ {2.22\times}$ & $ {1.86\times}$ & $ {2.23\times}$ & $ {2.22\times}$ \\ 

\midrule
Fast-dLLM+
& $ {2.69 \times}$ & $ {2.28\times}$ & $ {2.23\times}$ & $ {1.87\times}$ & $ {2.40\times}$ & $ {2.30\times}$ \\

\bottomrule

\end{tabular}

\vspace{-10pt}
\end{table*}

\subsection{SuffixDecoding}

SuffixDecoding~\cite{suffix_decoding} represents the state of the art non-parametric drafting method in speculative decoding. It maintains a suffix tree of tokens from previously generated text and the current prompt. If it finds an exact match of the most recent tokens generated, it ``speculates'' that the LLM will continue the sequence the same way it did in the past. SuffixDecoding only captures wins on the easy regions that exactly repeat the previous context. In contrast, \name{} adaptively determines which subsequences are easy and captures wins on easier segments that aren't just simply repeating prior context, gaining more wins.

In our evaluations, we find that SuffixDecoding provides negligible speedups over vanilla decoding on the workloads that we evaluate. Fundamentally, SuffixDecoding operates as a pattern-matcher rather than a predictive model; it yields the greatest efficiency gains when the LLM generates sequences identical to those in prior context, such as SQL schemas, boilerplate code, or repetitive reasoning loops. In agentic coding tasks (e.g., SWE-Bench), where an agent iteratively interacts with file systems and error logs, the method achieves high speedups by matching recurring token sequences. Conversely, in ``one-shot'' benchmarks such as HumanEval, and in mathematical reasoning chains where sequences are unique and lack frequent reflection, the suffix tree finds few matches.

\subsection{EAGLE-3}

\begin{figure*}[h]
    \centering
    \begin{subfigure}{0.32\textwidth}
        \centering
        \includegraphics[width=\linewidth]{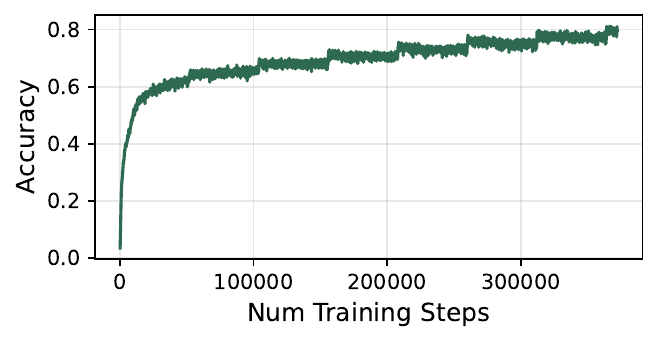}
        \caption{32B training.}
    \end{subfigure}
    \hfill %
    \begin{subfigure}{0.32\textwidth}
        \centering
        \includegraphics[width=\linewidth]{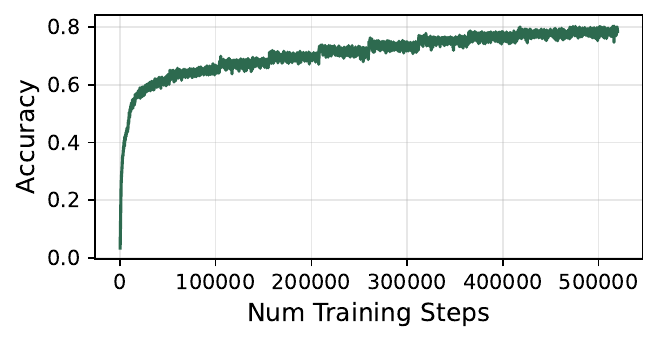} %
        \caption{14B training.}
    \end{subfigure}
    \hfill %
    \begin{subfigure}{0.32\textwidth}
        \centering
        \includegraphics[width=\linewidth]{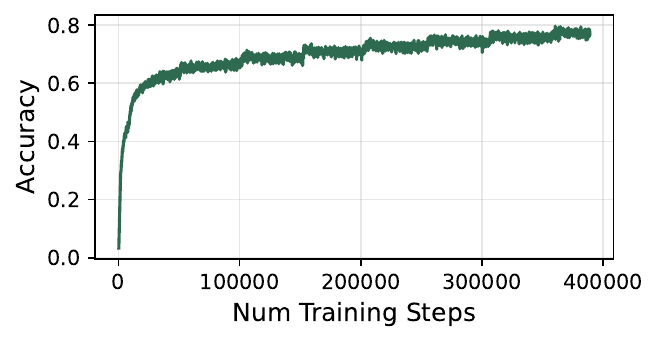} %
        \caption{7B training.}    
    \end{subfigure}
    \caption{EAGLE-3 training accuracy curves.}
    \label{fig:eagle3}
\end{figure*}

\textbf{EAGLE-3 training.} There are no public EAGLE-3 weights available for Qwen2.5. Thus, we conducted pretraining of EAGLE-3 weights using the ultrachat-200K~\cite{ultrachat} text corpus in SpecForge~\cite{specforge}. We adopted the default training hyperparameters in SpecForge and trained EAGLE-3 to match the target model's output until convergence.
We plot the moving average of the training accuracy in Fig.~\ref{fig:eagle3}.
During evaluation, we use the model checkpoints after five epochs of training ($\sim$260k training steps). We find that even though further training improves training-time accuracy, they have a negligible impact on the end-to-end speedup of EAGLE-3.
We open-source our EAGLE-3 model weights and training scripts.

\textbf{Draft token trees.} To ensure a fair and accurate representation of EAGLE-3’s peak performance~\cite{eagle3,eagle3_redhat}, we evaluate EAGLE-3 in SGLang v0.5.6.post2~\cite{sglang} with a draft token tree, with 50-60 tokens in each tree (following the best default hyperparameter configurations reported in the EAGLE-3 paper~\cite{eagle3}, roughly matching $N_{max}$ in \name{}).
For completeness, we additionally evaluate EAGLE-3 configured to propose a single chain of 3-20 draft tokens (rather than a tree). This variant consistently underperforms the full draft-tree version of EAGLE-3, and we therefore omit its results from the evaluation.
Similar to the AR drafter (Fig.~\ref{fig:theoretical_analysis}), adopting a fixed speculation length of 50-60 tokens in a single chain of draft tokens resulted in a significant slowdown compared to vanilla decoding due to the speculation overhead in autoregressive drafting.
We note that \name{}'s one-step generation is compatible with token trees, which can be realized via a custom attention mask~\cite{spiffy}. We leave this extension as future work.

\section{(Failed) Explorations of Alternative Design Choices and Limitations}
\label{sec:failed_explorations}

\textbf{Dynamic speculation length schemes.} Initially, borrowing on ideas in network congestion control~\cite{cc}, we tried to use the acceptance rate of previous rounds as an indication of token ``easiness'' in the current round. However, we found that the notion of easiness appears to be highly local and not well correlated across rounds. This finding prompted us to look forward (ahead into the future, i.e., the current tokens being speculated), not backward. 

\textbf{Reusing previous drafts.} In speculative decoding, all tokens following the first rejection are discarded. However, we observe that in many workloads -- particularly reasoning chains of thoughts -- rejections are often minor corrections (e.g., changing ``thus'' to ``therefore'', as can be observed in the example trajectory in Fig.~\ref{fig:failfast_trajectory}), while the subsequent tokens remain of high utility. In traditional speculative decoding, where speculation lengths are relatively short (3–16 tokens), reusing previous drafts offers marginal utility. In contrast, the aggressive speculation length expansion in \name{} makes reuse more beneficial; for example, if out of 60 proposed tokens, we accepted the first 19 tokens and corrected the 20th token, the remaining 40 tokens could be directly plugged back in if the 20th token was the only token that had to be corrected.
Nevertheless, because the drafting process in \name{} is inherently low-latency, the marginal speedups gained from reuse are often secondary.
When these cases do occur, they represent an opportunistic performance gain. For a specific example in the MATH dataset (Fig.~\ref{fig:failfast_trajectory}), we implemented a preliminary mechanism for reusing drafts: after each speculation round, we determine if the suffix of the current proposal exists within the tokens rejected in the previous round. If a match is found, the new proposal is appended with the tokens following that suffix.
For instance, if the rejected tokens from the last round are `xefghij' and the current draft is `abcdefg', our new proposal will be `abcdefghij'.
For this specific query, reuse provided a $\sim$15\% speedup over \name{}, reaching an impressive end-to-end speedup of 8.8$\times$ over vanilla decoding. 
However, since most rejections in other queries involve more than simple one-token corrections, the average speedup across the dataset is only $\sim$2\%, as the gains are diluted by rounds with fewer reuse opportunities. Consequently, we omit this technique from our primary results.

\textbf{Data-matched comparisons of parametric drafters.} We acknowledge that the training data used for our dLLM drafter and our EAGLE baseline do not exactly match. Our dLLM drafter, \texttt{Fast\_dLLM\_v2\_1.5B}, was adapted from \texttt{Qwen2.5-1.5B-Instruct} via a block-wise diffusion training process~\cite{fast_dllm_v2}. As a result, the difference in their performance might be partially attributed to this factor (although the comparisons with AR-drafters are indeed data-matched). Due to the significant amount of resources needed for a rigorous controlled experiment of different drafting methods, where training needs to be data-matched, parameter-matched, and FLOP-matched~\cite{roger_mamba_empirical_study,mamba2}, we leave it as future work and use this paper as a first step in demonstrating dLLM drafter's potential.

\section{Visualization of Example Trajectory}
\label{sec:traj_visualization}

\begin{figure*}[h]
    \centering
    \includegraphics[width=0.99\textwidth]{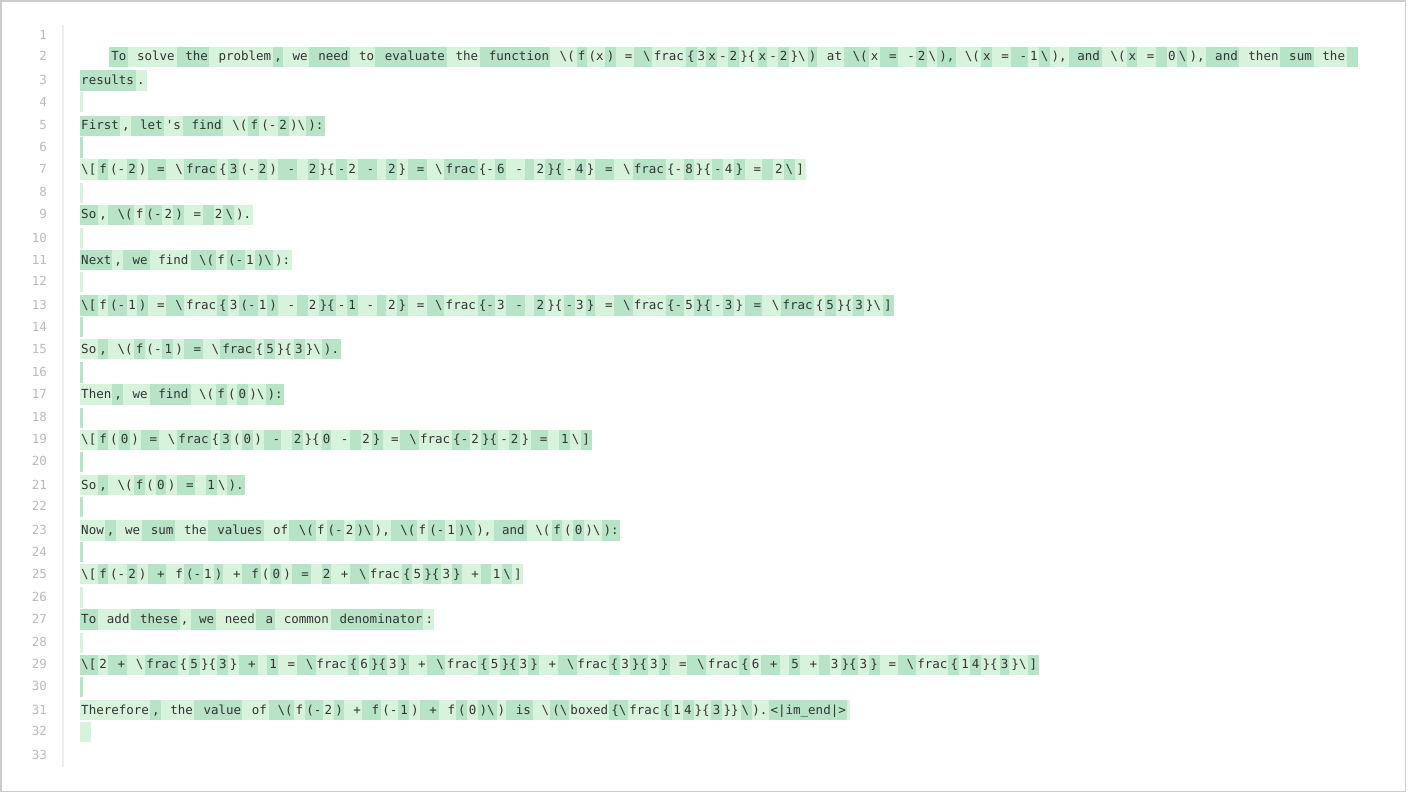}
    \caption{Ground truth trajectory.}
    \label{fig:accepted_trajectory}
\end{figure*}

In this section, we visualize the speculative decoding trajectory of an example query (question 2 from the MATH dataset). Individual tokens are distinguished by alternating background colors. Accepted tokens are shown as standard text. Rejected draft tokens are marked with a strikethrough and reduced opacity. Target-generated tokens (either corrections applied after a rejection or bonus tokens appended after a fully accepted draft) are highlighted with a solid border. This visualization illustrates the precise behavior of different speculative decoding baselines. The question is as follows:

\fbox{
  \begin{minipage}{0.95\linewidth}
    If $f(x) = \frac{3x-2}{x-2}$, what is the value of 
    $f(-2) + f(-1) + f(0)$? Express your answer as a common fraction.
  \end{minipage}
}

\begin{center}
\includegraphics[width=\linewidth]{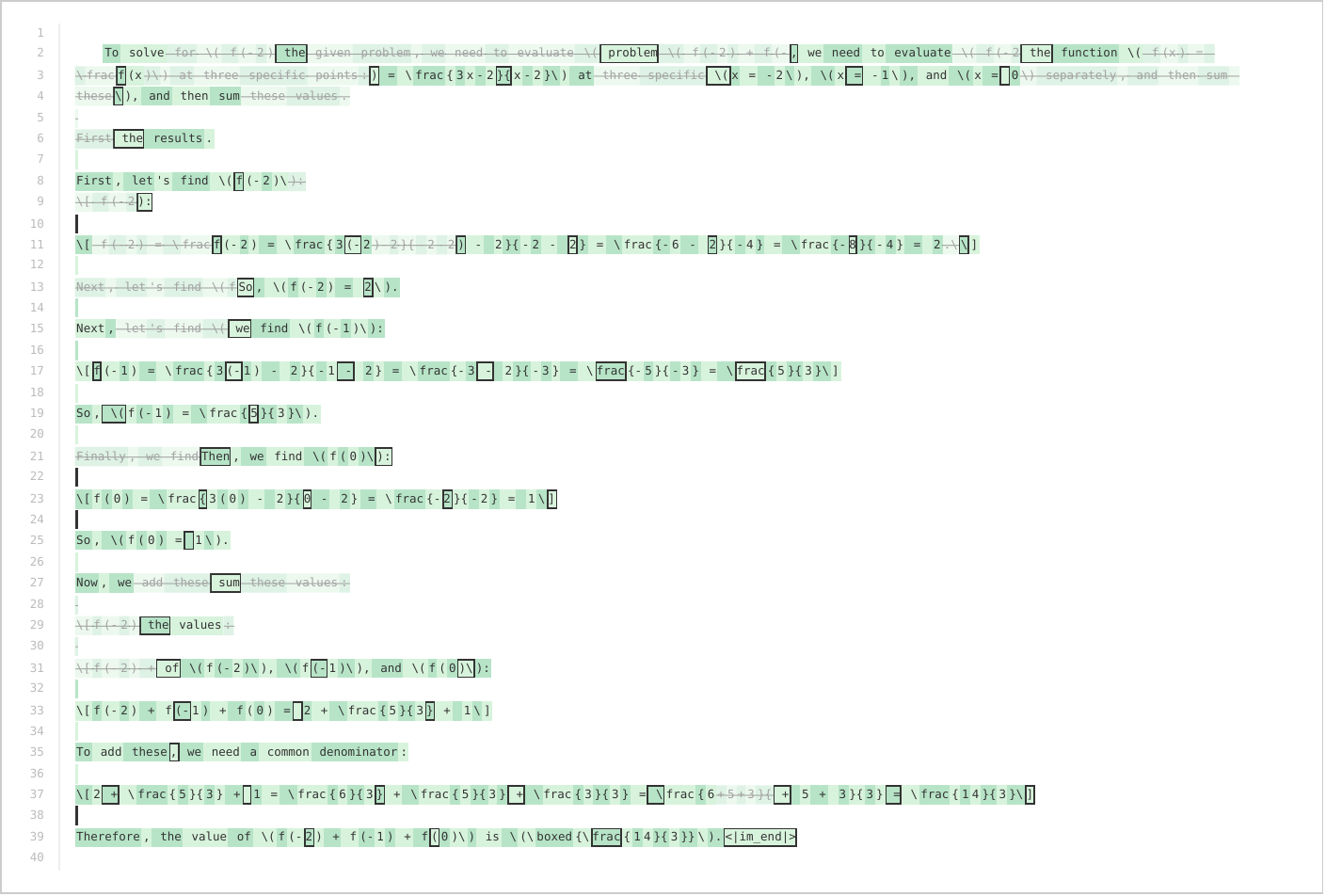}
\captionof{figure}{Autoregressive drafter's trajectory.}
\label{fig:sd_trajectory}
\end{center}

\begin{center}
\includegraphics[width=\linewidth]{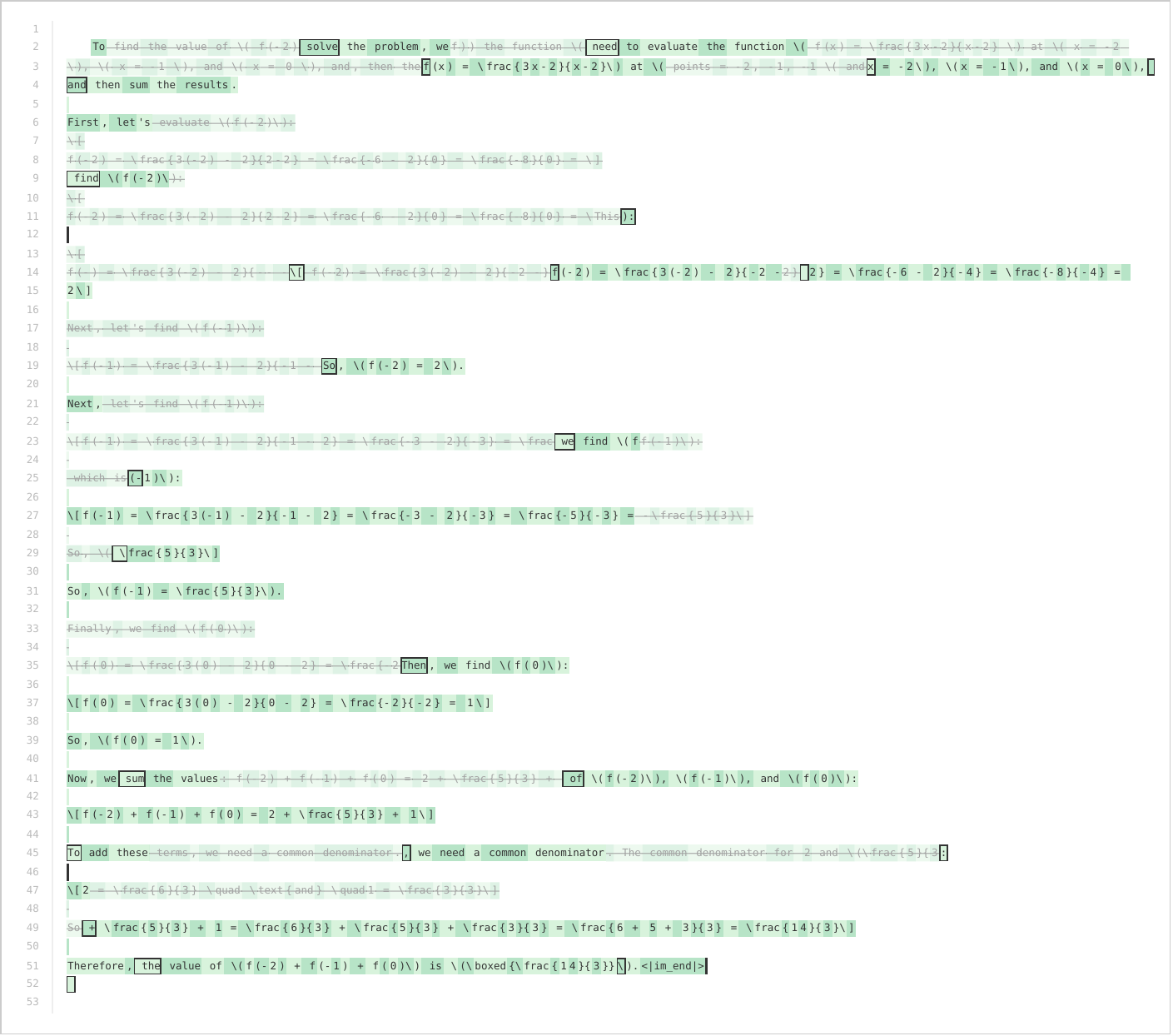}
\captionof{figure}{\name{}'s trajectory.}  %
\label{fig:failfast_trajectory}
\end{center}

In Fig.~\ref{fig:accepted_trajectory}, we first show the ground-truth trajectory as a reference. Fig.~\ref{fig:sd_trajectory} shows the trajectory of an autoregressive drafter with a static speculation length of 8 tokens. In many segments of the trajectory (e.g., line 11, 17, and 31-39), there are consecutive rounds where the acceptance rate is near-perfect; however, the target model is frequently invoked for verification in these segments, resulting in excessive latency overhead.

Finally, in Fig.~\ref{fig:failfast_trajectory}, we demonstrate the dynamic speculation length of \name{} in action. 
The main win from \name{}'s aggressive speculation length expansion comes from these rounds:
\squishlist
    \item Round 4 (line 3), 18/30 accepted (7 drafter forward passes in this round)
    \item Round 5 (line 3), 20/20 accepted (6 drafter forward passes in this round)
    \item Round 10 (line 14), 18/20 accepted (6 drafter forward passes in this round)
    \item Round 11 (line 14), 28/60 accepted (11 drafter forward passes in this round)
    \item Round 14 (line 38), 48/60 accepted (12 drafter forward passes in this round)
    \item Round 15 (lines 38-40), 23/60 accepted (11 drafter forward passes in this round)
    \item Round 16 (lines 42-48), 60/60 accepted (11 drafter forward passes in this round)
    \item Round 18 (lines 48-52), 50/50 accepted (9 drafter forward passes in this round)
    \item Round 22 (line 55), 60/60 accepted (12 drafter forward passes in this round)
    \item Round 23 (line 60), 30/30 accepted (7 drafter forward passes in this round)
\squishend

We note that the number of actual forward passes deviates from the intuition in Fig.~\ref{fig:intuition} due to a number of reasons: We do one-pass generation following Fast-dLLM's default small block size of 8, so generating 10 tokens might span across three small blocks, which requires three forward passes even though we are doing one-pass generation for each small block; occasionally, we need extra forward passes on a full block (e.g., 32 tokens) to populate the KV cache of prior drafted tokens~\cite{fast_dllm_v2}.